\documentclass[letterpaper]{article} 
\usepackage{aaai24}  
\usepackage{times}  
\usepackage{helvet}  
\usepackage{courier}  
\usepackage[hyphens]{url}  
\usepackage{graphicx} 
\usepackage{natbib}  
\usepackage{caption} 
\usepackage{algorithm}
\usepackage{algorithmic}
\usepackage{booktabs}
\usepackage{amsmath}
\usepackage{multirow}

\usepackage{xcolor}

\usepackage{newfloat}
\usepackage{listings}
\DeclareCaptionStyle{ruled}{labelfont=normalfont,labelsep=colon,strut=off} 
\floatstyle{ruled}
\newfloat{listing}{tb}{lst}{}
\floatname{listing}{Listing}
\title{DiffSynth: Latent In-Iteration Deflickering for Realistic Video Synthesis}
\author{
    Zhongjie Duan\textsuperscript{\rm 1},
    Lizhou You\textsuperscript{\rm 2},
    Chengyu Wang\textsuperscript{\rm 3},
    Cen Chen\textsuperscript{\rm 1},
    Ziheng Wu\textsuperscript{\rm 3},\\
    Weining Qian\textsuperscript{\rm 1},
    Jun Huang\textsuperscript{\rm 3}
}
\affiliations{
    \textsuperscript{\rm 1}School of Data Science and Engineering, East China Normal University\\
    \textsuperscript{\rm 2}School of Information, Xiamen University\\
    \textsuperscript{\rm 3}Alibaba Group\\

    \{zjduan,cenchen, wnqian\}@stu.ecnu.edu.cn\\
    youlizhou@xmu.edu.cn\\
    \{chengyu.wcy, ziheng.wzh, huangjun.hj\}@alibaba-inc.com
}

\usepackage{bibentry}

\begin{document}

\nocopyright
\maketitle

\begin{abstract}
In recent years, diffusion models have emerged as the most powerful approach in image synthesis. However, applying these models directly to video synthesis presents challenges, as it often leads to noticeable flickering contents. Although recently proposed zero-shot methods can alleviate flicker to some extent, we still struggle to generate coherent videos. In this paper, we propose DiffSynth, a novel approach that aims to convert image synthesis pipelines to video synthesis pipelines. DiffSynth consists of two key components: a latent in-iteration deflickering framework and a video deflickering algorithm. The latent in-iteration deflickering framework applies video deflickering to the latent space of diffusion models, effectively preventing flicker accumulation in intermediate steps. Additionally, we propose a video deflickering algorithm, named patch blending algorithm, that remaps objects in different frames and blends them together to enhance video consistency. One of the notable advantages of DiffSynth is its general applicability to various video synthesis tasks, including text-guided video stylization, fashion video synthesis, image-guided video stylization, video restoring, and 3D rendering. In the task of text-guided video stylization, we make it possible to synthesize high-quality videos without cherry-picking. The experimental results demonstrate the effectiveness of DiffSynth. All videos can be viewed on our project page\footnote{\url{https://anonymous456852.github.io/}}. Source codes will also be released\footnote{\url{https://github.com/alibaba/EasyNLP/tree/master/diffusion}}.
\end{abstract}

\section{Introduction}

In recent years, diffusion models have achieved remarkable success in the field of image synthesis, surpassing Generative Adversarial Networks (GANs) \cite{dhariwal2021diffusion}. In open-source communities, Stable Diffusion \cite{rombach2022high} has emerged as the most popular diffusion model, and fine-tuned models based on Stable Diffusion have achieved astonishing success in various artistic styles. Furthermore, several research breakthroughs have improved the capabilities of Stable Diffusion \cite{gal2022image, zhang2023adding, hulora, ruiz2023dreambooth, meng2021sdedit}. The achievements of Stable Diffusion have established it as the mainstream approach of image synthesis.

In our work, we further investigate the capabilities of diffusion models in video synthesis. It is well-known that directly applying image synthesis to each frame leads to significant flickering \cite{yu2021generating}. To address this problem, researchers have proposed solutions from various perspectives. One end-to-end solution is to pre-train a new video synthesis model using video datasets \cite{blattmann2023align}. Yet, there are no large-scale and high-resolution video datasets available to train video diffusion models, and the computational resources required for training are enormous. Furthermore, most existing techniques based on Stable Diffusion cannot be directly applied to newly trained models, making it difficult to control the generated results of such models. Therefore, we focus on one-shot and zero-shot methods, aiming to transfer existing image synthesis models to video synthesis with minimal or even no training. In recent years, several methods for video synthesis have been proposed and gained popularity. For example, Tune-A-Video \cite{wu2022tune} achieves video editing by fine-tuning the textual part of the model. Text2Video-Zero \cite{khachatryan2023text2video} restricts the content of adjacent frames using cross-frame attention. However, these methods are difficult to completely eliminate flicker in videos. Our aim is to design a more effective approach to fully eliminate flicker in videos synthesized by diffusion models.

To address the challenges mentioned above, we propose a novel approach named DiffSynth. Specifically, we design a latent in-iteration deflickering framework, with the aim of removing flicker during the intermediate iterations of video synthesis. 
In our approach, the video-level deflickering is applied to the decoded videos in the latent space. After that, the videos are encoded back from the latent deflickered representations.
Hence, we can effectively prevent the accumulation of flicker in denoising steps. 
For video deflickering, we design a patch blending algorithm to ensure the performance. By remapping different frames to the same frame, we can obtain the appearance features of the same object in different frames. The remapping operator is implemented based on patch matching \cite{barnes2009patchmatch}, which can estimate the nearest-neighbor field between two frames. Next, we blend the results to obtain a video with consistent contents. This algorithm can either eliminate high-frequency flicker using a sliding window, or thoroughly remove all flicker by blending all frames. Note that the latter has high computational complexity. Hence, we specifically propose a low time-complexity approximation algorithm, making our approach effective for long video synthesis.

DiffSynth is compatible with most models based on Stable Diffusion. Leveraging the research achievements in image synthesis, we have designed pipelines for multiple downstream tasks, including text-guided video stylization, fashion video synthesis, image-guided video stylization, video restoring, and 3D rendering. These video synthesis pipelines are all transferred from image synthesis pipelines, and we provide hyperparameter templates for these application scenarios to facilitate their usage. To validate the effectiveness of our method, we conducted extensive experiments in two application scenarios. Without any cherry-picking, we are able to generate coherent and realistic videos. Unsurprisingly, DiffSynth comprehensively outperforms existing baseline methods in quantitative metrics and user studies. We summarize the contributions of this paper as follows:
\begin{itemize}
    \item We propose DiffSynth, a novel approach for coherent and realistic video synthesis.
    \item We devise a latent in-iteration deflickering framework, which applies video-level deflickering to the latent space of diffusion models, avoiding the accumulation of flicker during the iterative process.
    \item Based on patch matching, we propose a patch blending algorithm that can significantly eliminate flicker in videos synthesized by diffusion models.
    \item We design several video synthesis pipelines utilizing DiffSynth for various tasks and demonstrate the superiority of our method in extensive experiments.
\end{itemize}

\section{Related Work}

\subsection{Diffusion Models}

Diffusion models \cite{song2019generative, nichol2021improved, sohl2015deep} are a kind of generative model that decomposes the image synthesis process into a sequential application of denoising models. Unlike GAN-based models \cite{goodfellow2020generative, huang2017stacked}, diffusion models do not rely on adversarial training and are generally easier to train. Latent Diffusion \cite{rombach2022high} introduces the concept of converting images from the pixel space to the latent space, making it possible to train the model on limited computational resources. Based on Latent Diffusion, Stable Diffusion becomes the most popular and powerful model in research communities. Recent studies have further enhanced the impressive capabilities of Stable Diffusion in various ways. LoRA (Low-Rank Adaptation) \cite{hulora} reduces computational resource requirements for fine-tuning by incorporating low-rank matrices into the model. To enhance the controllability of images generated by Stable Diffusion, ControlNet \cite{zhang2023adding} uses zero convolution to inject additional conditional information into the model, controlling the model generating objects in specific shapes, characters in specific poses, etc. Textual Inversion \cite{gal2022image} can synthesize a specific object by adding a new word to the vocabulary and training with a few images. Dreambooth \cite{ruiz2023dreambooth} further improves the model's ability to generate specific objects. Our proposed method is seamlessly compatible with these approaches, enhancing its capabilities in video synthesis.

\subsection{Diffusion-based Video Synthesis}

Inspired by the remarkable success of  diffusion models,
researchers endeavored to transfer diffusion models to video synthesis tasks, such as text-to-video synthesis, video style transfer and video editing. 
For example, Make-A-Video \cite{singer2022make} and VideoLDM \cite{blattmann2023align} extend a text-to-image diffusion model to a text-to-video diffusion model by incorporating temporal blocks. Gen-1 \cite{esser2023structure} utilizes a similar temporal architecture to transfer the video style. Besides pre-training a large-scale video model, recent studies focus on synthesizing videos using image models. Tune-A-Video \cite{wu2022tune} only fine-tunes several modules using input video, enabling video editing according to the given prompts. Some zero-shot methods, including FateZero \cite{qi2023fatezero}, Pix2Video \cite{ceylan2023pix2video}, and Text2Video-Zero \cite{khachatryan2023text2video}, have explored the possibility of synthesizing videos without additional training. These studies collectively demonstrate the feasibility of applying diffusion models to video synthesis tasks. Motivated by the findings of these studies, we propose a novel zero-shot approach, aiming to convert existing image synthesis pipelines to video synthesis pipelines expediently.

\section{Methodology}

In this section, we briefly review the diffusion models and then introduce our proposed approach.

\subsection{Preliminaries}

Generally, diffusion models include many different architectures \cite{feng2023ernie, saharia2022photorealistic, ramesh2022hierarchical}. In this paper, we focus mainly on Stable Diffusion \cite{rombach2022high}, which is the most popular open-source architecture. A typical Stable Diffusion model contains the following three components:
\begin{itemize}
    \item \textbf{Text Encoder}. A transformer-based language model in CLIP \cite{radford2021learning}, converting texts to text embeddings. The text embeddings are subsequently used in classifier-free guidance \cite{ho2021classifier}.
    \item \textbf{U-Net} \cite{ronneberger2015u}. A vision model $\epsilon$ with self-attention \cite{vaswani2017attention}, cross-attention, and residual connections \cite{he2016deep}. This model is trained to denoise images in the latent space.
    \item \textbf{VAE} \cite{2013Auto}. A model consists of an encoder $\mathcal E$ and a decoder $\mathcal D$, where the encoder converts images to latent tensors, and the decoder reconstructs images according to latent tensors.
\end{itemize}

Both the diffusion process and its reverse process (i.e., the generation process) are conducted in latent space. There are $T+1$ steps with $\{0,1,\dots,T\}$ representing different levels of noise. In the generation process, we start with a noise tensor $x_{T}$ sampled from a Gaussian distribution, and then denoise it stepwise. The iterative formula of each step $t$ is
\begin{equation}
    \small
    x_{t-1} = \sqrt{\alpha_{t-1}} \left(\frac{x_t-\sqrt{1-\alpha_t}\ \epsilon(x_t)}{\sqrt{\alpha_t}}\right) + \sqrt{1-\alpha_{t-1}}\ \epsilon(x_t),
    \label{equation:DDIM}
\end{equation}
where $\alpha_t$ is the hyperparameter describing how much noise it contains in step $t$. After the iterative process, the latent tensor $x_0$ is decoded to an image $X=\mathcal D(x_0)$. In addition to text-to-image synthesis, we can modify this process to implement other pipelines. For example, if we add noise to an image and then denoise from an intermediate step, we obtain an image-to-image pipeline for image editing.

To convert an image synthesis pipeline to a video synthesis pipeline, a naive approach is to generate each frame independently. However, most models based on Stable Diffusion are trained on text-image datasets and lack the ability to generate coherent video frames. After each iterative step, the difference between two adjacent frames becomes larger. The difference is accumulated during the generation process and finally results in irreparable flicker and inconsistency.

\subsection{Latent In-Iteration Deflickering}

As we mentioned above, the generation process of diffusion models is conducted in latent space, not pixel space. Inspired by deflickering methods designed for videos \cite{lei2020blind, lei2023blind}, we design a framework to apply deflickering to the latent space for better video synthesis.

For a video consisting of $n$ frames, at step $t$, we have $n$ latent tensors $\{x_t^1,x_t^2,\dots,x_t^n\}$ corresponding to each frame. If we directly compute $\{x_{t-1}^1,x_{t-1}^2,\dots,x_{t-1}^n\}$ using formula (\ref{equation:DDIM}), these latent tensors will become more inconsistent because each tensor is computed independently. To visualize the latent tensors, we first skip to the final step to calculate the estimation of $\{x_0^1,x_0^2,\dots,x_0^n\}$:
\begin{equation}
    \hat x_0^i = \frac{x_t^i-\sqrt{1-\alpha_t}\ \epsilon(x_t^i)}{\sqrt{\alpha_t}}.
    \label{equation:deflicker_start}
\end{equation}
Then, we decode $\{\hat x_0^1,\hat x_0^2,\dots,\hat x_0^n\}$ to images using the decoder component of VAE:
\begin{equation}
    \hat X^i=\mathcal D(\hat x_0^i).
\end{equation}
Theoretically, $\{\hat X^1,\hat X^2,\dots,\hat X^n\}$ represent the frames when we denoise $\{x_t^1,x_t^2,\dots,x_t^n\}$ along with a straight line directed by $\{\epsilon(x_t^1),\epsilon(x_t^2),\dots,\epsilon(x_t^n)\}$. We employ a video-level deflickering method $\mathcal F$ to make the video coherent, i.e.,
\begin{equation}
    \{\overline X^1,\overline X^2,\dots,\overline X^n\}=\mathcal F\{\hat X^1,\hat X^2,\dots,\hat X^n\}.
\end{equation}
Next, we encode the processed frames into the latent space:
\begin{equation}
    \overline x_0^i=\mathcal E\left(\overline X^i\right).
    \label{equation:vae_encode}
\end{equation}
To synthesize the video frames $\{\overline X^1,\overline X^2,\dots,\overline X^n\}$, the predicted noise at step $t$ should be:
\begin{equation}
    \overline \epsilon(x_t^i)=\frac{x_t^i-\sqrt{\alpha_t}\ \overline x_0^i}{\sqrt{1-\alpha_t}}.
\end{equation}
Thus, we obtain the reconstructed iterative formula as:
\begin{equation}
    x_{t-1}^i=\sqrt{\alpha_{t-1}}\ \overline x_0^i+\sqrt{1-\alpha_{t-1}}\ \overline \epsilon(x_t^i).
    \label{equation:deflicker_end}
\end{equation}
The pipeline (\ref{equation:deflicker_start}-\ref{equation:deflicker_end}) makes it possible to apply existing deflickering methods to latent tensors. At each denoising step, we can keep the difference between frames controllable and avoid the accumulation of flicker.

\begin{table*}
\centering
\tabcolsep=2pt
\renewcommand\arraystretch{1.3}
\vspace{-1em}
\begin{tabular}{rrrrrrrr}
\toprule
\multicolumn{1}{r}{\makebox[0.07\textwidth][r]{Frame 1}} & 
\multicolumn{1}{r}{\makebox[0.12\textwidth][r]{Frame 2}} & 
\multicolumn{1}{r}{\makebox[0.12\textwidth][r]{Frame 3}} & 
\multicolumn{1}{r}{\makebox[0.12\textwidth][r]{Frame 4}} & 
\multicolumn{1}{r}{\makebox[0.12\textwidth][r]{Frame 5}} & 
\multicolumn{1}{r}{\makebox[0.12\textwidth][r]{Frame 6}} & 
\multicolumn{1}{r}{\makebox[0.12\textwidth][r]{Frame 7}} & 
\multicolumn{1}{r}{\makebox[0.12\textwidth][r]{Frame 8}} \\ 
\hline
\multicolumn{1}{r}{$\hat X^1$} & 
\multicolumn{1}{r}{$\hat X^2$} & 
\multicolumn{1}{r}{$\hat X^3$} & 
\multicolumn{1}{r}{$\hat X^4$} & 
\multicolumn{1}{r}{$\hat X^5$} & 
\multicolumn{1}{r}{$\hat X^6$} & 
\multicolumn{1}{r}{$\hat X^7$} & 
\multicolumn{1}{r}{$\hat X^8$} \\ 
& 
\multicolumn{1}{r}{$[\hat X^{1\to 2}]_1$} &
& 
\multicolumn{1}{r}{$[\hat X^{3\to 4}]_1$} &
& 
\multicolumn{1}{r}{$[\hat X^{5\to 6}]_1$} &
& 
\multicolumn{1}{r}{$[\hat X^{7\to 8}]_1$} \\
&
& 
\multicolumn{2}{r}{$[\hat X^{1\to 4}]_2\oplus [\hat X^{2\to 4}]_1$} &
&
& 
\multicolumn{2}{r}{$[\hat X^{5\to 8}]_2\oplus [\hat X^{6\to 8}]_1$} \\
& 
& 
& 
& 
\multicolumn{4}{r}{$[\hat X^{1\to 8}]_3\oplus [\hat X^{2\to 8}]_2\oplus [\hat X^{3\to 8}]_2\oplus [\hat X^{4\to 8}]_1$} \\
\bottomrule
\end{tabular}
\caption{An example of a remapping table in the patch blending algorithm.}
\label{table:remapping_table}
\vspace{-1em}
\end{table*}

\subsection{Patch Blending Algorithm}

Another problem to be addressed is how to design the video-level deflickering method $\mathcal F$ mentioned above. In a video synthesis task without any reference, we can only employ blind video deflickering methods \cite{lei2020blind, lei2023blind}, thus it is difficult to keep the video fluent. In many video synthesis tasks (e.g., style transferring and video editing), we have the original video for reference. Hence, we mainly focus on these tasks in this work and propose the following deflickering algorithm.

Assuming that we have obtained the synthesized frames $\{\hat X^1,\hat X^2,\dots,\hat X^n\}$ in one denoising step, we aim to make the video smoother with reference to the original video $\{X^1,X^2,\dots,X^n\}$. For an object in frame $\hat X^i$, it may also occur in another frame $\hat X^j$, thus we intend to remap the corresponding areas in $\hat X^j$ to $\hat X^i$ and then blend them together. The blended frame will show the consistent information of both $\hat X^i$ and $\hat X^j$ if the remapping result is accurate. Although recent studies \cite{kirillov2023segment, han2022mat} have achieved impressive success in object segmentation and tracking, it is still difficult to achieve pixel-level accuracy. We have also considered optical flow \cite{teed2020raft}, but we find that optical flow is typically not accurate when the interval of two frames is large. Finally, we decide to utilize a patch matching algorithm \cite{barnes2009patchmatch}, which is an effective algorithm to estimate the correspondence between two frames. In this algorithm, the two frames $X^i$ and $X^j$ are divided into some overlapping patches. We first compute a \textit{nearest neighbor field} (NNF) that represents the matched patches and then reconstruct $\hat X^i$ using $\hat X^j$. Finally, we blend the reconstructed frames with the synthesized frames.

As an efficient implementation, we only remap and blend in a small sliding window. 
Yet, in the worst case, we need to remap and blend all frames together, which requires $\mathcal O(n^2)$ times of NNF estimation. The high time complexity is the main pitfall that prevents this algorithm from application. To improve computational efficiency, we propose an $\mathcal O(n\log n)$ approximate algorithm.

For convenience, we use a remapping operator $[j\to i]$ to denote the remapping process and use $\hat X^{j\to i}$ to denote the remapped results from $\hat X^j$ to $\hat X^i$. Note that the remapped and blended results could be remapped to another frame again, but multiple times of remapping and blending may result in non-negligible errors. We use a subscript number to denote the times of remapping and blending, i.e.,
\begin{equation}
    [\hat X^{j\to i}]_0=\hat X^i,
\end{equation}
\begin{equation}
    [\hat X^{j\to i}]_{u+1}=[\hat X^{j\to k}]_u[k\to i].
\end{equation}
The blending process is denoted as a blending operator $\oplus$. If we directly use the average, this operator is equivalent to ``+''. Note that the coefficient $\frac{1}{2}$ is not included in the blending operator. The final blended result should be divided by the number of frames. We can also train a lightweight network to implement this operator, in order to reduce the errors in the following approximate calculation. We leave this component for future work.

We use a remapping table to store some intermediate variables. An example of a remapping table is presented in Table \ref{table:remapping_table}. First, we store each synthesized frame in the first row. Then in the $k$-th iteration and the $i\cdot 2^k$-th column, we blend the frames in the $(i\cdot 2^k-2^{k-1})$-th column and then remap it to the $i\cdot 2^k$-th frame. For example, see the $3$rd row and the $4$th column in Table \ref{table:remapping_table}. It is calculated as follows:
\begin{equation}
\begin{aligned}
    &\left(
    [\hat X^{1\to 2}]_1 \oplus \hat X^2
    \right)
    [2\to 4]\\
    =&
    \left([\hat X^{1\to 2}]_1 [2\to 4]\right)
    \oplus
    \left(\hat X^2 [2\to 4]\right)\\
    =&
    [\hat X^{1\to 4}]_2\oplus [\hat X^{2\to 4}]_1
\end{aligned}
\end{equation}
With another array storing the ``$\oplus$'' sum of each column, we can calculate the whole table within $\mathcal O(n)$ time complexity, and the memory complexity is also $\mathcal O(n)$.

Once we obtain the remapping table, we can quickly calculate the smoothed frame $\overline X^i=\frac{1}{n}\oplus_{j=1}^n \hat X^{j\to i}$. For instance, when $i=6$, we have:
\begin{equation}
\begin{aligned}
    &\oplus_{j=1}^6 \hat X^{j\to 6}\\
    \approx&
    \left([\hat X^{1\to 4}]_2\oplus [\hat X^{2\to 4}]_1\oplus [\hat X^{3\to 4}]_1 \oplus \hat X^4\right)[4\to 6]\\
    &\oplus \left([\hat X^{5\to 6}]_1 \oplus \hat X^6\right)\\
    =&[\hat X^{1\to 6}]_3\oplus [\hat X^{2\to 6}]_2\oplus [\hat X^{3\to 6}]_2 \oplus [\hat X^{4\to 6}]_1\\
    &\oplus [\hat X^{5\to 6}]_1 \oplus \hat X^6
\end{aligned}
\label{equation:left_query}
\end{equation}
Similarly, we reverse the sequence and calculate the remaining part:
\begin{equation}
    \oplus_{j=7}^8 \hat X^{j\to 6}
    \approx[\hat X^{7\to 6}]_1\oplus [\hat X^{8\to 6}]_2
\label{equation:right_query}
\end{equation}
As we mentioned above, the ``$\oplus$'' sum of each column has been stored in another array, thus we can calculate the estimation of $\oplus_{j=1}^n \hat X^{j\to i}$ within $\mathcal O(\log n)$ time.

Taking into account the accuracy of the estimated ``$\oplus$'' sum, we can easily prove that the maximum number of remapping and blending operations is $\mathcal O(\log n)$. In equations (\ref{equation:left_query}) and (\ref{equation:right_query}), it is apparent that the frames close to the $i$-th frame exhibit lower error compared to those located further away. Consequently, this algorithm achieves both speed and numerical stability.

\subsection{Other Modifications}

To convert an image synthesis pipeline to a video synthesis pipeline, we further modify the following details.

\begin{itemize}
    \item \textbf{Fixed noise}. When we synthesize images, sampling from the same Gaussian noise leads to the same image if we leave other settings fixed. In video synthesis, the frames in a video are expected to be similar; thus we synthesize each frame from the same Gaussian noise. In some downstream tasks, some information from the input video is supposed to be retrained in the edited video, thus we add the same Gaussian noise to each frame.
    \item \textbf{Cross-frame attention}. If we want to generate an image similar to a reference image, we can concatenate our image with the reference image and synthesize our image in an in-painting pipeline. This is a widely used trick to control the generated content. However, the model will draw unexpected components near the seam line, because it tends to combine the two images into one complete image. Essentially, the information from the reference images is passed to our image mainly by self-attention \cite{vaswani2017attention}. Thus, we change self-attention layers to cross-frame attention layers. In the denoising process of the $i$-th frame, we concatenate $x_1$, $x_{i-1}$, $x_i$ and $x_{i+1}$ together in the self-attention layers. If a ControlNet is used in the pipeline, we also convert self-attention layers of ControlNet to cross-frame attention layers.
    \item \textbf{Adaptive resolution}. Today, most videos on the social network are in the shape of rectangles, and their resolution is usually high. However, most diffusion models are trained to synthesize images in the shape of $512\times 512$ squares. Considering the architecture of U-Net, we can easily change the shape and increase the resolution according to the downstream tasks. In our experiments, we surprisingly find that higher resolution sometimes leads to more fluent video frames. The reason is that the implicit patches represent more fine-grained information when the resolution is higher.
    \item \textbf{Deterministic sampling of VAE}. In equation (\ref{equation:vae_encode}) and image-to-image pipelines, the VAE encoder is called to encode images to latent space. In fact, the output of VAE is a Gaussian distribution, not a deterministic tensor. To maintain the consistency of different frames, we use deterministic sampling instead of Gaussian sampling (i.e., setting the standard deviation to zero).
    \item \textbf{Memory-efficient attention}. Cross-frame attention greatly increases the time consumed to synthesize a video, and the memory required is up to $\mathcal O(N^2)$, where $N$ is the number of implicit patches. Existing studies have shown that the attention mechanism could be implemented in low memory \cite{rabe2021self}. Therefore, we employ flash attention \cite{dao2022flashattention} to implement the attention mechanism, which is capable of significantly improving computational efficiency and reducing the memory required.
    \item \textbf{Smoothed ControlNet annotator}. When a ControlNet model is used to control the content, an annotator is employed for processing the original frames, but we observed that some annotators may cause flickering control signal. For example, the OpenPose \cite{cao2017realtime} annotator can cause tic of the limbs if the frames are not clear enough. To overcome this pitfall, we use Savitzky-Golay smoothing filters \cite{press1990savitzky} to smooth the coordinates of the key points detected.
\end{itemize}

\section{Experiments}

To demonstrate the effectiveness of our approach, we conducted the following experiments, including text-guided video stylization and fashion video synthesis. We compared our approach with several state-of-the-art approaches and evaluated the synthesized videos using both quantitive metrics and user study.

\subsection{Experimental Settings}

\begin{table}
\centering
\tabcolsep=2.8pt
\begin{footnotesize}
\begin{tabular}{l|cc}
\toprule
                                  & Text-guided    & Fashion        \\
                                  & video          & video          \\
                                  & stylization    & synthesis      \\ \hline
Frame height                      & 512            & 768            \\
Frame width                       & 960            & 512            \\
Denoising steps                   & 20             & 10             \\
\multirow{2}{*}{ControlNet scale} & 1.0 (Depth)    & 0.3 (Depth)    \\
                                  & 1.0 (SoftEdge) & 0.6 (OpenPose) \\
CFG scale                         & 7.5            & 7.5            \\
Deflickering window size          & $\infty$       & 7              \\
Deflickering frequency            & 5              & 1              \\ \bottomrule
\end{tabular}
\end{footnotesize}
\caption{The hyperparameters in the experiments.}
\label{table:parameters}
\vspace{-1.5em}
\end{table}

\noindent\textbf{Text-Guided Video Stylization.}
In this task, we design a pipeline to transfer the style of videos according to given prompts. The dataset, consisting of 100 high-resolution videos, is collected from a community\footnote{https://pixabay.com/}. Each video is cut into 3 to 5 seconds, including at most 150 frames. We manually write prompts for each video\footnote{\url{https://github.com/ECNU-CILAB/Pixabay100}}. The pipeline is composed of a popular customized model in open-source communities\footnote{https://civitai.com/models/4384/dreamshaper} and two ControlNet models. We use Depth\footnote{https://huggingface.co/lllyasviel/control\_v11f1p\_sd15\_depth} and SoftEdge\footnote{https://huggingface.co/lllyasviel/control\_v11p\_sd15\_softedge} to provide structure guidance. The information in the original video is only delivered to the pipeline by ControlNet, thus the color is completely ignored. After the final step, we add an additional deflickering step to reduce the flicker generated in the last four steps and set the window size to 121. We further improve the contrast ratio and sharpen the frames slightly to improve video quality.

\begin{figure}
  \includegraphics[width=1.0\linewidth]{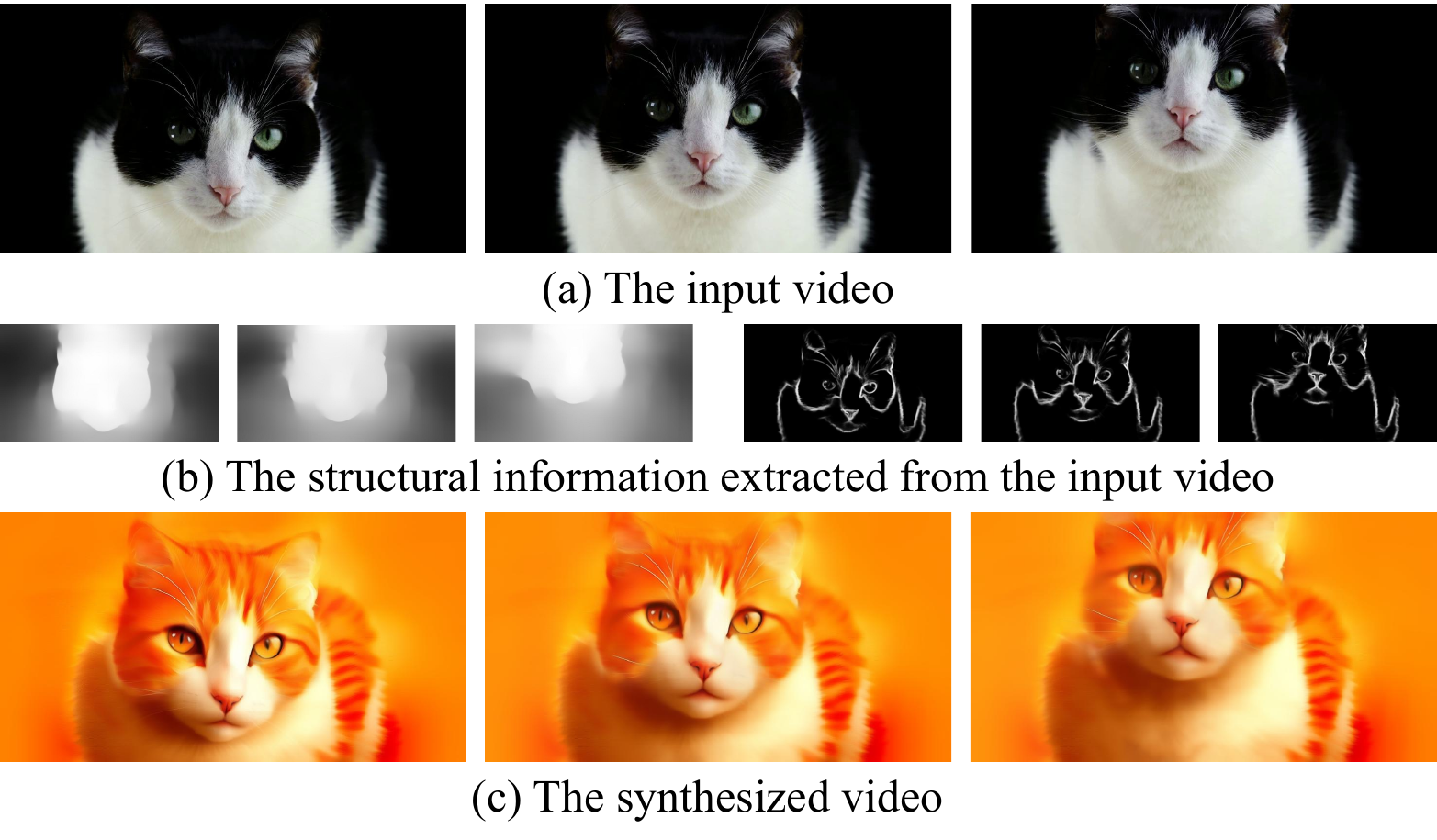}
  \vspace{-1.5em}
  \caption{An example of text-guided video stylization. The prompt in this example is ``an orange and white cat''.}
  \label{fig:StyleTransfer}
\end{figure}

\noindent\textbf{Fashion Video Synthesis.}
The second task is to customize virtual fashion models and synthesize fashion videos on e-commerce platforms. The dataset used in this experiment is a fashion video dataset \cite{zablotskaia2019dwnet}, which contains hundreds of videos. There is a fashion model with fashion clothes in each video. We randomly select 10 source videos from the dataset and fine-tune Stable Diffusion 1.5\footnote{https://huggingface.co/runwayml/stable-diffusion-v1-5} on each video, respectively. The fine-tuned models have learned the appearance of each fashion model; in other words, one fine-tuned diffusion model represents a virtual fashion model. We randomly select the other 10 target videos, then extract the pose sequence using OpenPose \cite{cao2017realtime}, and let the 10 virtual fashion models imitate the pose in the other 10 target videos. In this pipeline, we use OpenPose ControlNet\footnote{https://huggingface.co/lllyasviel/control\_v11p\_sd15\_openpose} to control the pose of models. We notice that the pose extracted by OpenPose suffers from slight tic, although we use Savitzky-Golay smoothing filters \cite{press1990savitzky}. To further smoothen the synthesized videos, we use Depth ControlNet to stabilize the motion. Finally, we obtain $10\times 10$ realistic synthesized videos.

\begin{figure}
  \includegraphics[width=1.0\linewidth]{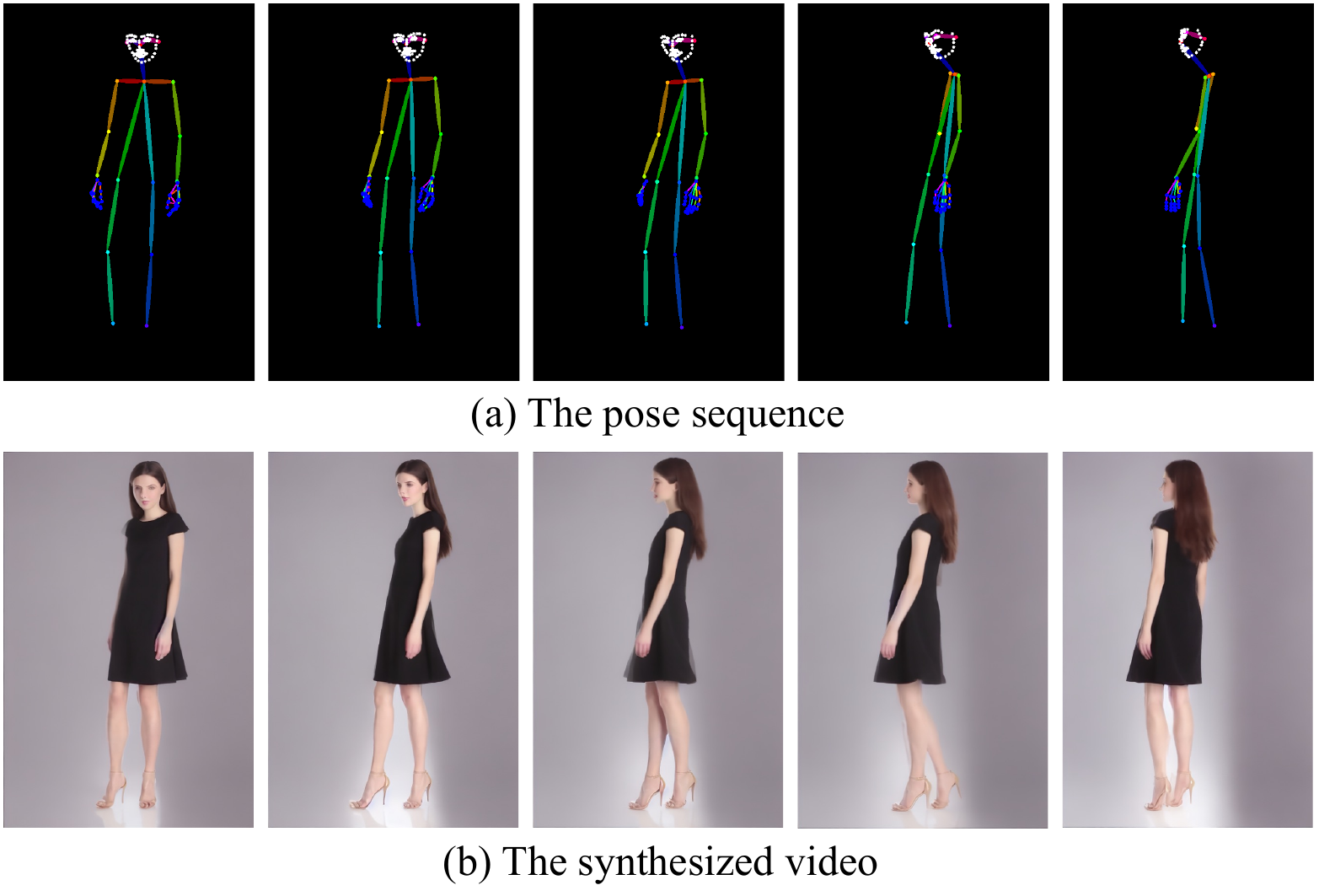}
  \vspace{-1.5em}
  \caption{An example of fashion video synthesis.}
  \vspace{-1em}
  \label{fig:fashhion_video_synthesis}
\end{figure}

\noindent\textbf{Hyperparameters.}
The hyperparameters of the above two tasks are presented in Table \ref{table:parameters}. These parameters are tuned by human experts. The size of the deflickering window is $\infty$, denoting that we use the fast remapping and blending algorithm to blend all frames. The deflickering frequency denotes how frequently we apply the deflickering algorithm. For example, when the number of denoising steps is 20 and the deflickering frequency is 5, we apply the deflickering algorithm at the 1st, 6th, 11th, and 16th steps. To make the experiments reproducible, we synthesize each video with the same random seed, without any cherry-picking.

\begin{table}
\centering
\small
\tabcolsep=4.0pt
\begin{tabular}{l|c|cc}
\toprule
                & Content                  & Prompt                  & Content                      \\
                & Consistency              & Similarity              & Aesthetics                    \\
                & {\scriptsize (Pixel-MSE $\downarrow$)} & {\scriptsize (CLIP Score $\uparrow$)} & {\scriptsize (Aesthetic Score $\uparrow$)} \\ \hline
FateZero        & -                        & 22.32                   & 5.44                         \\
Pix2Video       & 1203.04                  & 25.29                   & 5.20                         \\
Text2Video-Zero & 342.85                   & 23.86                   & 5.13                         \\
DiffSynth       & \textbf{54.97}           & 24.63                   & 5.37                         \\ \bottomrule
\end{tabular}
\caption{Quantitative results in text-guided video stylization.}
\vspace{-.5em}
\label{table:metric_stylization}
\end{table}

\begin{table}
\centering
\small
\tabcolsep=4.0pt
\begin{tabular}{l|c|cc}
\toprule
                      & Content                  & Appearance         & Pose                    \\
                      & Consistency              & Similarity         & Error                   \\
                      & {\scriptsize (Pixel-MSE $\downarrow$)} & {\scriptsize (FID $\downarrow$)} & {\scriptsize (Pose-MSE $\downarrow$)} \\ \hline
DreamPose             & 100.44                   & 75.43              & 0.63                    \\
DiffSynth             & \textbf{24.13}           & 63.02              & 0.48                    \\ \bottomrule
\end{tabular}
\caption{Quantitative results in fashion video synthesis.}
\vspace{-1.5em}
\label{table:metric_fashion}
\end{table}

\begin{table}[t]
\centering
\small
\tabcolsep=3.2pt
\begin{tabular}{l|cccc}
\toprule
Approach   & FateZero & Pix2Video & Text2Video-Zero & DiffSynth      \\ \hline
Percentage & 24.43    & 7.65      & 12.07           & \textbf{55.85} \\ \bottomrule
\end{tabular}
\caption{Percentage of videos selected as best in user study.}
\vspace{-.5em}
\label{table:user_study}
\end{table}

\subsection{Quantitive Results}

We evaluate the quality of synthesized videos, with results presented in Table \ref{table:metric_stylization} and Table \ref{table:metric_fashion}. In text-guided video stylization, we compare our approach with several baseline approaches, including FateZero \cite{qi2023fatezero}, Pix2Video \cite{ceylan2023pix2video}, and Text2Video-Zero \cite{khachatryan2023text2video}. In fashion video synthesis, we compare it with DreamPose. Following Pix2Video, we calculate Pixel-MSE to evaluate the consistency of the content. Note that the official code of FateZero does not support the resolution except $512\times 512$, so we cannot calculate its Pixel-MSE. From this metric perspective, DiffSynth clearly outperforms other approaches. It shows that our method can generate smoother videos than others. In addition, we computed other metrics to evaluate the quality of each frame. In text-guided video stylization, we employ CLIP score \cite{radford2021learning} to measure the relevance between the synthesized video and the prompt, and use the aesthetic score \cite{schuhmann2022laion} to assess the aesthetics of frames. FateZero can only make minor changes to the frames, failing to generate videos that match the textual description. Pix2Video is excessively focused on textual information, resulting in incoherent frames. Text2Video-Zero performed slightly better than the aforementioned methods but still lags behind our approach. In fashion video synthesis, we employ the Fréchet Inception Distance (FID) \cite{heusel2017gans} to measure the appearance similarity between synthesized videos and source videos, and utilize Pose-MSE (Mean Squared Error of the keypoint distances recognized by OpenPose \cite{cao2017realtime}) to assess the pose error between synthesized videos and target videos. Our method exceeds DreamPose in all metrics. These experimental results demonstrate the effectiveness of DiffSynth.

\subsection{User Study}

Some studies \cite{blattmann2023align, yang2023rerender} have pointed out that conventional metrics are sometimes not feasible. We invite 20 participants and conduct a user study. We ask each participant to select the best results based on video consistancy, text-video similarity, and aesthetics. The average results of the 100 videos in text-guided video stylization are presented in Table \ref{table:user_study}. Unsurprisingly, most of the participants think that the videos synthesized by DiffSynth look better than others.

\begin{figure}
  \includegraphics[width=1.0\linewidth]{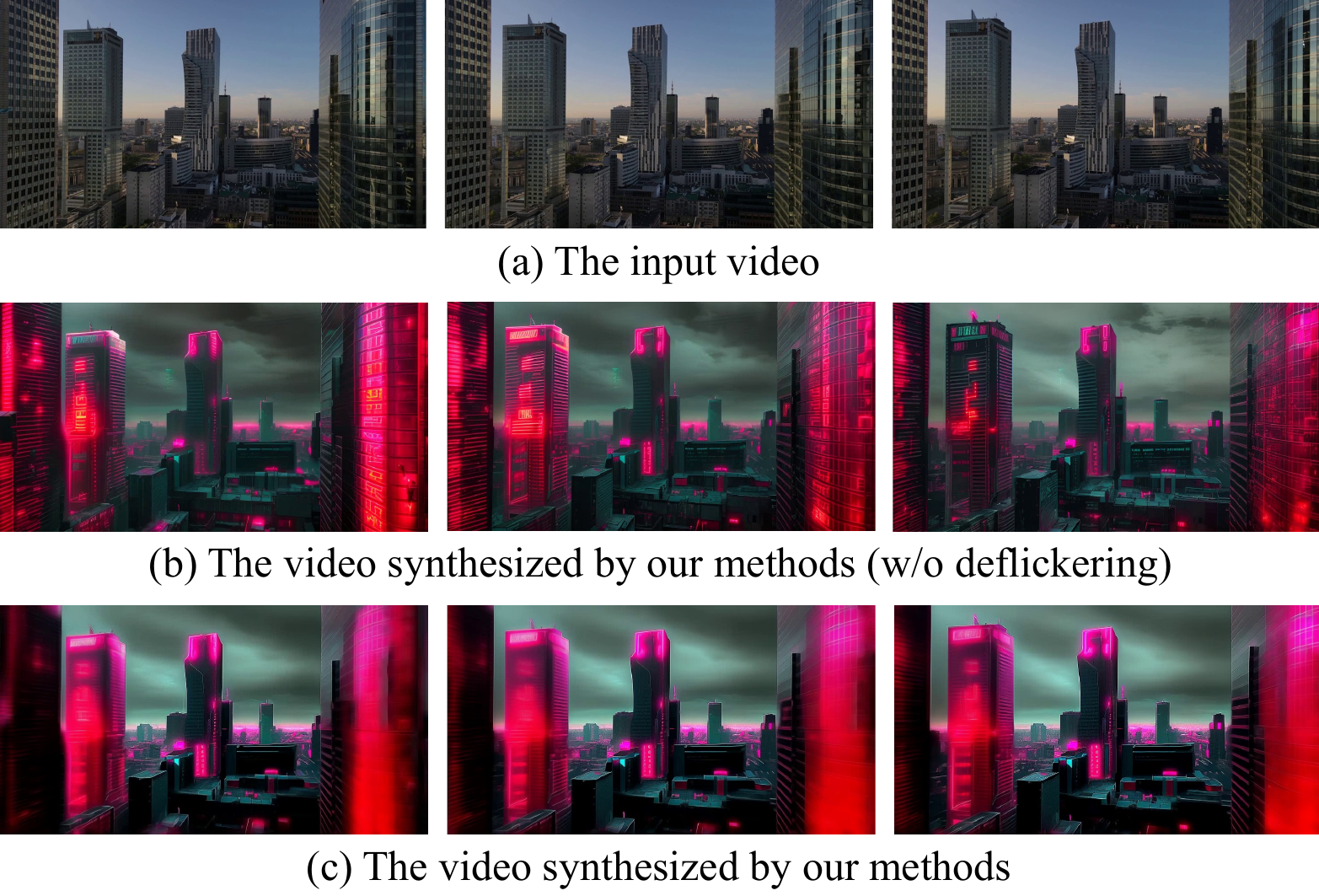}
  \caption{An example of ablation study. The prompt of this example is ``cyberpunk, city, red neon light''.}
  \vspace{-.5em}
  \label{fig:ablation}
\end{figure}

\subsection{Ablation Study}

To evaluate the effectiveness of our proposed deflickering algorithm (i.e., the patch blending algorithm), we conduct an ablation experiment. Figure \ref{fig:ablation} illustrates an example of our ablation study. For more examples, please refer to our project page. When the deflickering algorithm is disabled, it is evident that the video exhibits inconsistencies in various aspects, including the brightness of the sky, the lights on the buildings, and the texts at the central building. When the deflickering algorithm is enabled, these objects are more aligned, resulting in more coherent videos.

\section{Other Applications}

Besides the above two tasks, we also design several fancy pipelines in the following scenarios. These video synthesis pipelines are all transferred from image synthesis pipelines.

\subsection{Image-Guided Video Stylization}

\begin{figure}
  \includegraphics[width=1.0\linewidth]{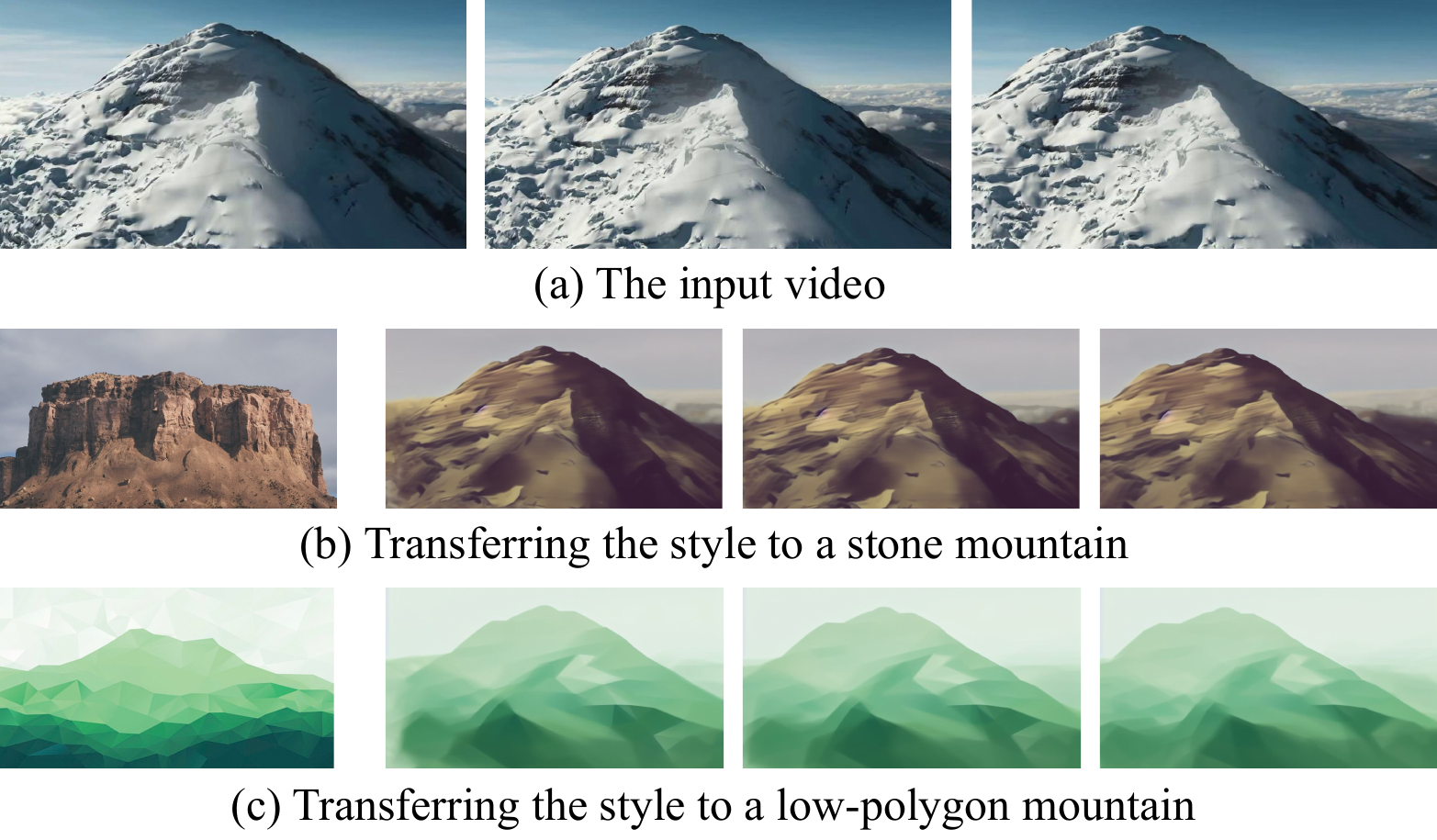}
  \caption{Examples of image-guided video stylization.}
  \label{fig:image_StyleTransfer}
\end{figure}

In the above stylization pipeline, the frames are synthesized according to input prompts. Practically, the synthesized videos are greatly influenced by the prompts, and carefully tuned prompts can improve the quality of videos. Prompt engineering has become an interesting but challenging task \cite{witteveen2022investigating}. To intuitively guide the model generating videos, we employ a ControlNet model\footnote{https://huggingface.co/lllyasviel/control\_v11e\_sd15\_shuffle} to further enable the image guiding mechanism. As shown in Figure \ref{fig:image_StyleTransfer}, this pipeline can transfer the style of a content video according to the style of an image. Compared to the above pipeline, this pipeline does not require well-written prompts. Therefore, it is easy to use for creators who are not familiar with diffusion models.

\subsection{Video Restoring}

\begin{figure}
  \includegraphics[width=1.0\linewidth]{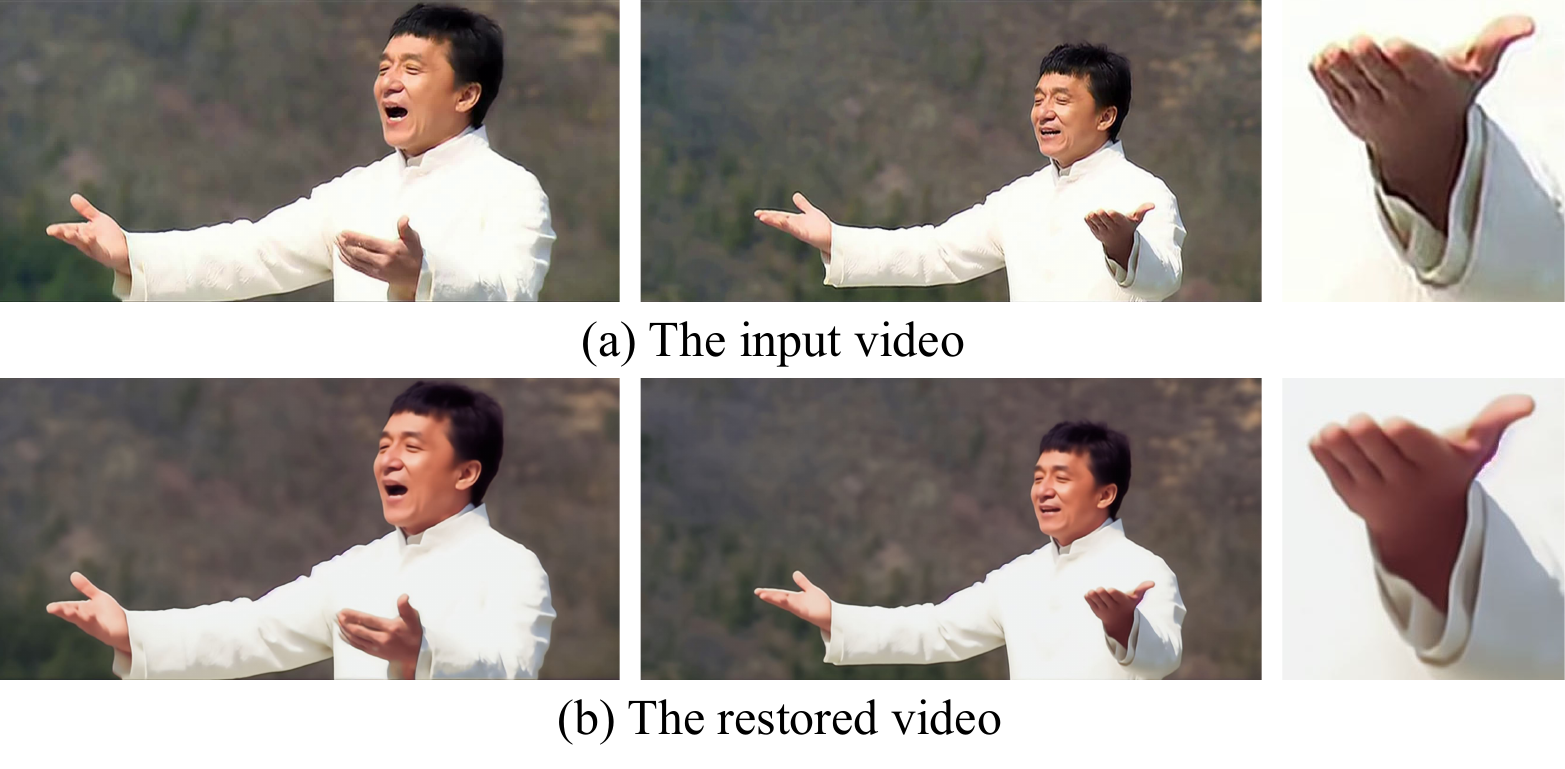}
  \caption{An example of video restoring.}
  \label{fig:video_restoring}
  \vspace{-1em}
\end{figure}

In earlier years, due to the immaturity of photography, some old videos only have very low resolution and may have hard-to-repair noise because of repeated compression during distribution. Restoration of these videos requires a lot of costs. One method of restoration using deep learning is super-resolution, but the defects such as noise in the original video are also retained in the restored video. We decided to use our video processing method to restore these videos. Combined with Tile ControlNet\footnote{https://huggingface.co/lllyasviel/control\_v11f1e\_sd15\_tile}, the Stable Diffusion model can ignore the defects in old videos and redraw noisy video frames. Specifically, we first increase the resolution of the original video using the super-resolution method and then map each frame to the latent space. After adding noise to the intermediate steps, we subsequently use Tile ControlNet for denoising. Unlike in the other application scenarios, we do not generate each frame from scratch because we tend to preserve as much information as possible in the original video, with only limited modifications to the details. Figure \ref{fig:video_restoring} shows an example. We can see that our pipeline is capable of restoring the details of historic videos.

\subsection{3D Rendering}

\begin{figure}
  \includegraphics[width=1.0\linewidth]{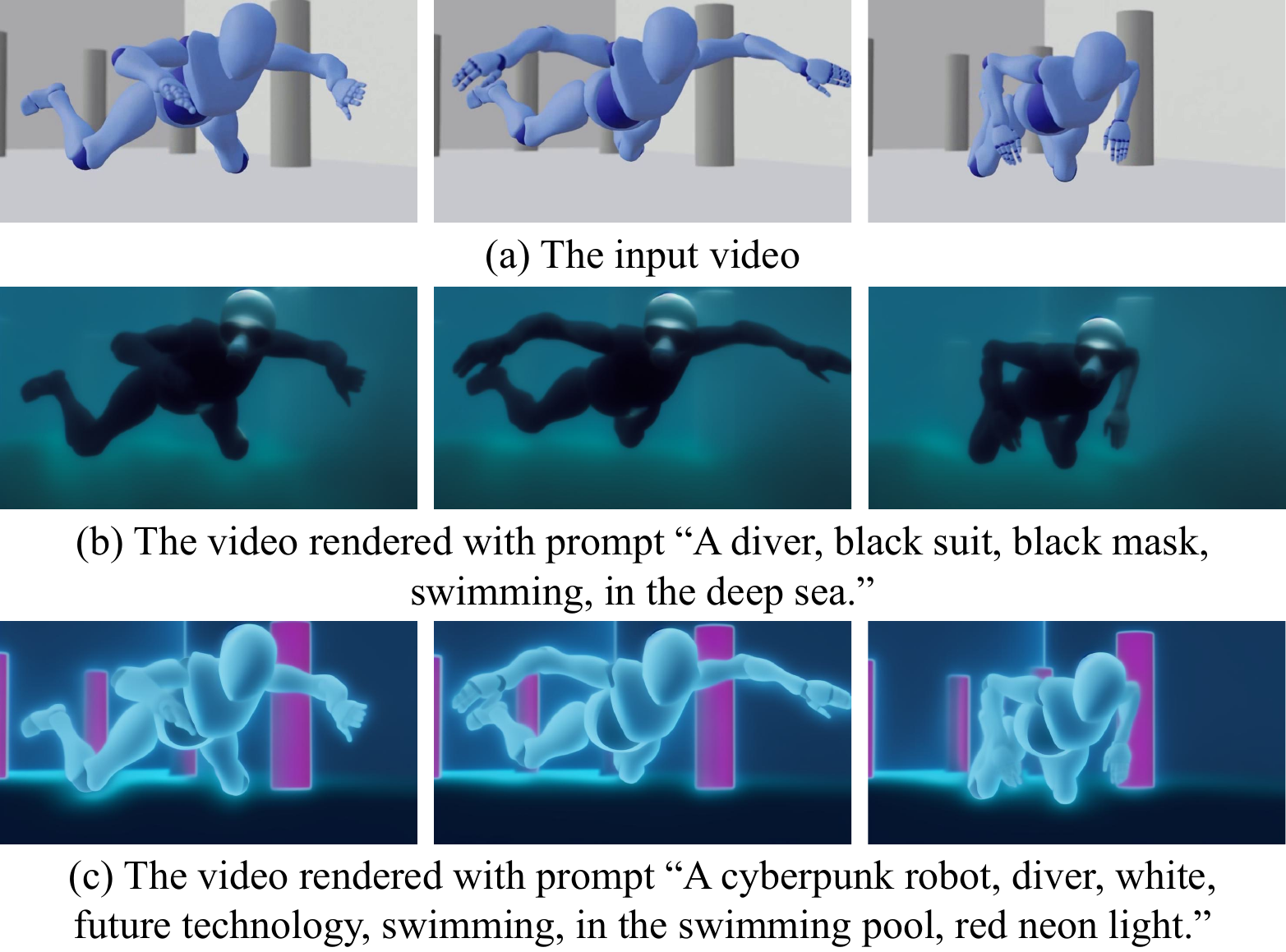}
  \caption{Examples of 3D rendering.}
  \label{fig:3d_rendering}
    \vspace{-1em}
\end{figure}

In the media industry, creators have to draw maps for a 3D object to render a video. Utilizing our approach, we can directly render a video automatically. We first extract some necessary information from the unrendered video frames to represent the structure of frames. There are no unique answers to the definition of what structural information is, and we take reference from Gen-1 \cite{esser2023structure} to extract depth information. In addition, we also use the SoftEdge information if necessary. We then design a ControlNet-based pipeline to render the video. An example is presented in Figure \ref{fig:3d_rendering}. Our pipeline can transform the 3D gray object to a realistic object, and it only requires creators to provide unrendered videos and prompts.
Therefore, our work has the capacity of benefiting the media industry by video designs.

\section{Conclusion and Future Work}

In this paper, we investigate the application of diffusion models to video synthesis. We propose the latent in-iteration deflickering approach, making it possible to apply existing video deflickering methods to the latent space, thereby avoiding flicker accumulation during the iterative process. We specifically design a deflickering algorithm based on patch matching for diffusion models. With this algorithm, we can synthesize coherent and realistic videos without any cherry-picking. We further show that our approach is applicable to various application scenarios. Comprehensive experimental results demonstrate the effectiveness of our method, outperforming previous methods.

Yet, our work still has a few limitations. Although the $\mathcal O(n\log n)$ time complexity can be achieved, the efficiency of our approach can be further improved for wider applications. For example, running on an NVIDIA A10 GPU, our program requires approximately 1 minute to synthesize 1 frame in fashion video synthesis. Furthermore, the blending operator can be improved to generate details better. We leave these problems for future work.

\bibliography{aaai24}

\end{document}